# No-Reference Image Contrast Assessment with Customized EfficientNet-B0

Javad Hassannataj Joloudari[1,2,3,*], Bita Mesbahzadeh[4], Omid Zare[5], Emrah Arslan[6], Roohallah Alizadehsani[7], Hossein Moosaei[8]

[1]Department of Computer Engineering, Faculty of Engineering, University of Birjand, Birjand, Iran, javad.hassannataj@birjand.ac.ir
[2]Department of Computer Engineering, Babol Branch, Islamic Azad University, Babol, Iran
[3]Department of Computer Engineering, Technical and Vocational University (TVU), Tehran, 4631964198, Iran
[4]Department of Industrial Engineering, K. N. Toosi University of Technology, Tehran, Iran, b.mesbahzadeh@gmail.com
[5]Department of Computer Science, University of Verona, Verona, Italy, omid.zare@univr.it
[6]Department of Computer Engineering, Faculty of Engineering, KTO Karatay University, Konya, Türkiye, emrahaslan2608@gmail.com
[7]Institute for Intelligent Systems Research and Innovation (IISRI), Deakin University, Waurn Ponds, Australia, r.alizadehsani@deakin.edu.au
[8]Department of Informatics, Faculty of Science, Jan Evangelista Purkyně University, Ústí nad Labem, Czech Republic hossein.moosaei@ujep.cz

*Corresponding author: javad.hassannataj@birjand.ac.ir

**Abstract.** Image contrast was a fundamental factor in visual perception and played a vital role in overall image quality. However, most no-reference image quality assessment NR-IQA models struggled to accurately evaluate contrast distortions under diverse real-world conditions. In this study, we proposed a deep learning-based framework for blind contrast quality assessment by customizing and fine-tuning three pre-trained architectures, EfficientNet-B0, ResNet18, and MobileNetV2, for perceptual Mean Opinion Score (MOS), along with an additional model built on a Siamese network, which indicated a limited ability to capture perceptual contrast distortions. Each model is modified with a contrast-aware regression head and trained end-to-end using targeted data augmentations on two benchmark datasets, CID2013 and CCID2014, containing synthetic and authentic contrast distortions. Performance is evaluated using Pearson Linear Correlation Coefficient (PLCC) and Spearman Rank Order Correlation Coefficient (SRCC), which assess the alignment between predicted and human-rated scores. Among these three models, our customized EfficientNet-B0 model achieved state-of-the-art performance with PLCC = 0.9286 and SRCC = 0.9178 on CCID2014 and PLCC = 0.9581 and SRCC = 0.9369 on CID2013, surpassing traditional methods and outperforming other deep baselines. These results highlighted the model's robustness and effectiveness in capturing perceptual contrast distortion. Overall, the proposed method demonstrated that contrast-aware adaptation of lightweight pre-trained networks can yield a high-performing, scalable solution for no-reference contrast quality assessment suitable for real-time and resource-constrained applications.

1. **Introduction**

Image quality assessment (IQA) has played a vital role in designing and optimizing image and video processing systems, with many algorithms developed recently. These methods were classified into full-reference (FR), reduced-reference (RR), and no-reference (NR) based on whether a distortion-free reference image was available [1]. Image distortions such as compression artifacts, additive noise, blur, and contrast degradation significantly affected visual perception and reduced the performance of downstream tasks like object recognition or medical image analysis [2, 3].

Although contrast distortion had a significant impact on perceived image quality, it has been largely overlooked in IQA research. Consequently, most existing no-reference IQA methods demonstrated poor performance when dealing with contrast-distorted images, as this type of distortion has been rarely addressed in previous studies. [4-6]. The importance of this issue can be stated as among the various aspects of image quality, contrast was a key cue through which the human visual system separated structure from background; recent reviews tied poor contrast to missed pathology and workflow delayed in radiology [7]. Clinical head-CT work further showed that deep learning reconstruction delivering a 15–20 % contrast-to-noise gain can measurably shift diagnostic confidence [8]. Beyond medicine, accurate contrast scoring supported low-light enhancement, segmentation, and autonomous navigation, where subtle tonal cues guided critical decisions [9]. Yet, collecting dense human ratings for every modality was impractical, reinforcing the demand for compact self-contained NR-IQA frameworks.

For traditional FR models, such as the Structural Similarity index (SSIM) [10] and Feature Similarity index (FSIM) [11] An incorporated contrast comparison function is explicitly included to reflect the human visual system's sensitivity to contrast masking. Thus, these models detected contrast-related degradations far more reliably than simpler error-based metrics like Peak Signal-to-Noise Ratio (PSNR) and Mean Squared Error (MSE) [12]. More advanced methods like Visual Information Fidelity (VIF) utilized natural scene statistics and multiscale wavelet domain models, showing superior correlation with subjective perception in contrast-sensitive scenarios [13].

Recently, IQA-RR models have been developed to assess contrast distortion using limited reference data. Reduced-Reference Image Quality model for Contrast-changed images (RIQMC) utilized entropy and the first four statistical moments, along with phase congruency, to produce contrast-sensitive quality scores [14]. QMC improved perceptual alignment by combining entropy and saliency features from both references and distorted images. The RCIQM model employed a hybrid framework that combined bottom-up analysis, through free energy modeling, with top-down evaluation via histogram comparison. While the bottom-up component captured structural variations in the visual content, the top-down strategy accounts for distributional change in luminance and contrast, enabling a more comprehensive assessment of image quality [15]. Results demonstrated these methods rival or surpass FR models, highlighting their utility in low-reference scenarios.

Contrast distortion in NR-IQA has been directly addressed by only a few models, beginning with traditional approaches such as the Contrast Enhanced Image Quality measure (CEIQ), which generated a histogram-equalized pseudo-reference and extracted SSIM, histogram entropy, and cross-entropy features, later regressed via SVR to estimate perceptual quality [4]. Other handcrafted pipelines, such as a robust PCA-based contrast quality estimator, model brightness, sharpness, and statistical contrast moments to handle both under- and over-enhanced images without a reference [16].

In more advanced learning-based models, Mahmoudpour et al. proposed a pseudo-reference selection network that generated multiple contrast-enhanced candidates, selected the most natural one via a trained classifier, and applied a full-reference model (e.g., SSIM) between it and the distorted input, effectively bridging NR and FR IQA for contrast [17]. General deep NR-IQA frameworks, such as Meta-learning based Image Quality Assessment (MetaIQA) [18] and SaTQA [19], can handle contrast distortions when trained on relevant datasets, but they were not specifically optimized for contrast degradation.

Contrast distortion remained a critical challenge in NR-IQA due to its strong perceptual impact and the lack of a reference image. Traditional approaches primarily relied on handcrafted features derived from natural scene statistics or histogram characteristics. For example, Fang et al. hypothesized that contrast distortion would temper the statistical regularities of natural images, leading to unnatural appearance that degraded perceived image quality, and designed an NSS-based NR model using statistical moment deviations and entropy measures [20]. Similarly, Zhu et al. introduced three statistical features: local contrast, histogram shape, and brightness, and employed SVR to predict the quality of contrast-distorted images [21].

Another notable method, CEIQ, generated a histogram equalized pseudo reference and measures similarity using SSIM, entropy, and cross entropy features, which are then fused for regression [4]. More advanced models have incorporated hybrid features combining information-theoretic and appearance attributes for improved prediction [21]. However, these handcrafted pipelines often failed to generalize to authentic distortions and struggled to capture high-level semantic cues. To overcome these limitations, deep learning-based models have emerged; for example, Hu et al. proposed a Convolutional Neural Network (CNN) framework pre-trained on synthetic contrast-distorted images [22] and fine-tuned on IQA datasets in an end-to-end manner [5]. The remaining gaps included the lack of large-scale contrast-specific datasets, insufficient robustness to diverse distortion levels, and limited integration of local contrast cues with global semantics.

To address these gaps, we leveraged a customized EfficientNet-B0 backbone and applied extensive data augmentation to improve generalization and perceptual sensitivity following recent trends in CNN-based IQA [19, 23]. To resolve these issues, this study proposed an NR-IQA model using a customized EfficientNet-B0 backbone. It reviewed traditional and deep learning-based methods before introducing a hybrid loss combining MOS regression and ranking to enhance sensitivity to contrast variations. Experiments on CCID2014 and CID2013 showed that a customized EfficientNet-B0 had superior accuracy and efficiency compared to ResNet-18 and MobileNetV2. The study also analyzed trade-offs, highlighted dataset limitations, and suggested future improvements for broader validation.

Efficient-Net's compounded scaling offers an appealing backbone for such lightweight yet expressive models. The B0 variant recently outperformed deeper networks while retaining the lowest memory footprint in practical vision tasks [24] However, its capacity for fine-grained contrast judgment has not been examined. our model achieved state-of-the-art correlation with subjective contrast ratings while remaining computationally efficient.

The main contributions of our study are as follows.

- This work introduced EfficientNet-B0 as an effective and lightweight backbone for NR-IQA.
- The model achieved a strong balance between accuracy and efficiency.
- It integrated MOS regression with PLCC/SRCC evaluation to ensure both predictive accuracy and perceptual alignment.
- Data augmentation is employed to enhance the model's robustness against diverse distortions.
- The customized EfficientNet-B0 achieved state-of-the-art performance with PLCC = 0.9286 and SRCC = 0.9178 on CCID2014, and PLCC = 0.9581 and SRCC = 0.9369 on CID2013 outperforming classical models and recent deep IQA models.

The remaining study was organized as follows: Section 2 reviewed prior FR, RR, and NR approaches, handcrafted and deep, with notes on Siamese designs and Efficient-Net. Section 3 detailed the methodology: datasets (CCID2014, CID2013), preprocessing, the proposed architecture, and training settings. Section 4 reported results and comparisons (PLCC/SRCC, ablations, and limitations), Section 5 discussed implications, Section 6 concluded with contributions and future work, and Section 7 listed declarations.

**2. Related Work**

In the related work, different strategies for assessing contrast distortion have been explored. Section 2.1 This section described FR IQA methods that were specifically designed to assess contrast distortions and improve alignment with human visual perception. Section 2.2 covered reduced-reference approaches that relied on dedicated databases and statistical features to capture change in contrast. Section 2.3 turned to handcrafted no-reference methods, where natural scene statistics, entropy measures, and histogram analysis are used without the need for reference images. Section 2.4 introduced deep learning–based solutions, including CNNs, meta-learning, and weakly supervised models. This section also included 2.4.1, which focused on Siamese architectures designed to improve perceptual alignment through ranking and attention mechanisms, and 2.4.2, which discussed Efficient-Net-based approaches that provided lightweight yet accurate models, recently adapted for contrast quality assessment. Section 2.5 The table compared image quality assessment methods (FR, RR, NR) and their approaches to evaluating contrast distortions.

*2.1 FR IQA Methods for Contrast Distortion*

FR IQA models have been developed specifically for contrast-distorted images. Patch-based Contrast Quality Index (PCQI) decomposed patches into mean, contrast, and structure, achieving high consistency with human opinion on CID2013/CCID2014 datasets [25]. Local Linear Contrast Model (LLCM) enhanced perceptual alignment by linearly combining luminance, contrast, and structural variations at the local level [26]. Finally, Full-Reference Contrast based on Singular Value Decomposition (FC-SVD) captured global contrast energy variations through singular value analysis, providing improved sensitivity to distortions [27].

*2.2. RR-IQA Methods for Contrast Distortion*

Efforts that exploited partial reference image data have produced tools specifically tailored to contrast distortion: Gu et al. Designed the CID2013 and CCID2014 databases and proposed RR-IQA techniques based on phase congruency and image histogram statistics. These methods enabled the assessment of contrast change using condensed reference descriptors [28]. Liu et al. Presented RCIQM, integrating free energy theory with histogram-based comparisons between contrast-altered and reference images, enabling perceptually adjusted quality estimation[29]. Even though they are effective when reference data is partially available, these methods fell short in no-reference settings [6].

*2.3. Handcrafted NR-IQA Methods*

Fang et al. [20] utilized deviations in natural scene statistics, such as moments and entropy, to estimate contrast degradation without a reference. NIQMC [28] was a training-free NR model based on local entropy and histogram uniformity comparison under the information maximization principle. CEIQ [4] created a histogram equalized pseudo reference and derived SSIM, entropy, and cross entropy features for regression-based contrast quality estimation. Zhu et al. [30] combined local contrast histogram shape and brightness features refined through a Just Noticeable Difference (JND) model, then projected onto MOS using SVR. MDM [15] used just three features, Minkowski distance and entropy, to deliver efficient and accurate NR-IQA for contrast distortion across datasets.

*2.4. Deep Learning–Based NR-IQA Methods*

Advances in deep learning have shifted approaches toward representation learning frameworks that implicitly encode contrast behavior. Hu et al. [5] Used a two-stage deep CNN pre-trained on synthetic contrast distorted data and fine-tuned for NR-IQA, explicitly addressing a gap in contrast distortion assessment via deep learning. RankIQA [31] Introduced a Siamese network trained on pairwise quality rankings from synthetically distorted images, which effectively transfers to single-image quality estimation. MetaIQA [18] Employed deep meta-learning to learn shared quality priors across multiple distortions, including contrast, achieving flexible adaptation to novel distortions. DeepFL-IQA [32] Proposed a weakly supervised feature learning framework using a large-scale IQA dataset and multitask learning, achieving better results than FR models on specific benchmarks.

### 2.4.1. Siamese Architectures

Siamese networks have been applied to image contrast assessment by learning relative perceptual rankings between references and distorted images via shared weight branches, as in RankIQA, which achieved strong results on TID2013 and LIVE [31]. The Gradient Siamese Network (GSN) enhanced contrast sensitivity through central differential convolution spatial attention and multi-level feature fusion, securing second place in NTIRE 2022 Full-Reference IQA[33]. Similarly, the Attention-based Siamese Difference Network (ASNA) integrated spatial channel attention with surrogate ranking loss to better align with human perceptual judgments in the contrast-related IQA challenge [34]. In our study, we integrated a pretrained model with a Siamese network; however, the approach demonstrated limited performance in detection and evaluation tasks.

### 2.4.2. EfficientNet for Perceptual Quality

The pursuit of practical deployment has led to the development of a lightweight yet accurate model where Transfer learning from image classification backbones has proven effective in IQA, for instance, Bianco et al. [35] and Song et al. [36] finetuned pre-trained networks achieved strong cross-dataset generalization. Among these backbones, EfficientNet, with its compound scaling strategy [24] has demonstrated state-of-the-art accuracy per parameter, making it a promising candidate for IQA tasks. Its adoption was gradually increasing, as seen in IE-IQA [37] and in the MRI [38] quality index proposed by Ramachandran et al. Building on this trend, recent studies have explored large-scale variants such as EfficientNet-B7, which was employed in BMC Imaging [37] for MRI quality assessment. Furthermore, an ensemble approach combining EfficientNet-B0 [39] and NASNet Mobile has been investigated in DeepEnsembling [40]. Despite these advancements, little prior work has tailored EfficientNet to contrast quality. Our customized EfficientNet-B0 filled this gap by integrating contrast-specific attention and ranking objectives for accurate, efficient blind contrast assessment. Our approach extended this line of research by customizing the EfficientNet-B0 architecture with augmented datasets, a model recognized for its superior parameter efficiency and scalable feature extraction. This architecture enabled more effective contrast assessment compared to previous methods.

### 2.5. Overview of Image Quality Assessment Models for Contrast Evaluation

Table 1 summarized the key FR, RR, and NR models applied to image contrast assessment, describing their core methodologies from handcrafted features to deep learning. The table highlighted how each approach estimated perceived contrast quality, illustrating the evolution from traditional error models to advanced neural networks. This overview emphasized the need for improved NR methods tailored specifically to contrast distortions.

Table 1. Models Proposed for Image Contrast Assessment.

| Reference | Category | Model | Methodology | Results (PLCC, SRCC) In CID2013 and CCID2014 |
|---|---|---|---|---|
| Yan et al., 2019 [4] | NR | CEIQ | Generate an enhanced image through histogram equalization and compute SSIM between the original and enhanced image as the first feature; compute entropy and cross-entropy between their histograms, and fuse 5 features via SVR. | CID2013: PLCC 0.952, SRCC 0.939 CCID2014: PLCC 0.949, SRCC 0.936 |
| Gu et al., 2015 [14] | RR | RIQMC | Combine 'similarity' (PC-based entropy) and 'comfort' (order statistics: mean, std, skewness, kurtosis) with linear fusion; requires one reduced reference number. | CCID2014: PLCC 0.8726, SRCC 0.8465 |
| Gu et al., 2017 [41] | NR | NIQMC | Local contrast via entropy on salient regions; global contrast via symmetric KL divergence between image histogram and uniform; combine local and global. | CID2013: PLCC 0.922, SRCC 0.901, CCID2014: PLCC 0.925, SRCC 0.904 |
| Fang et al., 2015 [20] | NR | Fang NR-IQA | NSS-based features tailored for contrast-altered images. | CID2013: PLCC 0.933, SRCC 0.919, CCID2014: PLCC 0.936, SRCC 0.919 |
| Gu et al., 2017 [28] | NR | MDM | Uses high-order Minkowski distance with power-law transformation and entropy (3 features) mapped to MOS via SVR. | CCID2014: PLCC 0.8719, SRCC 0.8368 |
| Wang et al., 2004 [42] | FR | SSIM | Measures luminance, contrast, and structural similarity between reference and test images. | CID2013: PLCC 0.498, SRCC 0.465, CCID2014: PLCC 0.681, SRCC 0.670 |

According to the referenced studies, each model showed clear limitations. SSIM required a pristine reference image and was not specifically designed for contrast distortions, leading to weak consistency with human perception on contrast-altered datasets. CEIQ demonstrated strong results on its training datasets but suffered from poor cross-database generalization, showing reduced reliability when applied to unseen data. RIQMC depended on reduced-reference information, which restricted its applicability in blind assessment scenarios, and its performance is limited to natural images with contrast enhancement rather than general distortions. NIQMC, while effective, was computationally expensive and thus less practical for large-scale or real-time applications. Fang NR-IQA relied heavily on supervised learning with support vector regression, making it data-dependent and sensitive to the choice and size of training datasets. Finally, MDM is tailored to contrast distortions using only a few handcrafted features, which constrained its generality and limited its ability to handle a broader range of image quality degradations.

## 3. Research methodology

In this research, convolutional neural network architectures, namely MobileNetV2, EfficientNet-B0, and ResNet, were customized and fine-tuned for the task of image contrast assessment. The overall methodology consists of four main stages: Data description, Data Preprocessing, Feature Extraction, and Fine-tuning & MOS Prediction. The research methodology for image contrast assessment is illustrated in Figure 1. This architecture ensured that the pre-trained models contributed strong general feature representations, while the customized dense layers adapted the model. Finally, a regression layer outputs MOS, representing the contrast quality of the image.

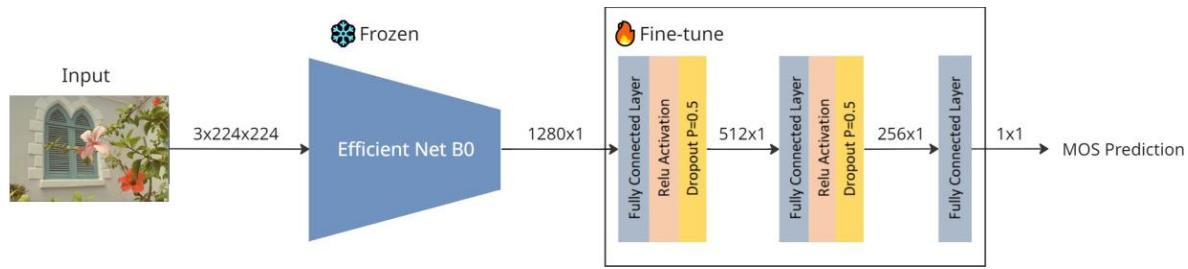

**Fig. 1** Model Architecture.

The pseudocode of the pre-trained model is presented below.

---

**Algorithm 1: Customized Deep Learning Model for contrast IQA**

Input:

- Training dataset (images + MOS scores)
- Pre-trained backbone model (EfficientNet-B0, ResNet18, MobileNetV2)

Output:
- Trained Contrast IQA model
1. Begin

2. Load dataset and apply preprocessing:
   - Resize images
   - Normalize pixel values
   - Split into training/validation

3. Initialize backbone with pre-trained weights (num_classes=0)

4. Freeze backbone layers to retain pre-trained features

5. Construct a custom regression head:
   - Linear [in-features → 512] + ReLU + Dropout(0.5)
   - Linear [512 → 256] + ReLU
   - Linear [256 → 1]

6. Attach the regression head to the backbone

7. Define loss function: Mean Squared Error (MSE)

8. Choose optimizer: AdamW (lr=1e-4)

9. For each training epoch, do:
   9.1 Forward pass: frozen backbone → regression head
   9.2 Compute MSE loss
   9.3 Backpropagate (only regression head)
   9.4 Update parameters
   9.5 Validate (PLCC, SROCC)
10. If validation plateaus: unfreeze backbone and fine-tune with smaller lr
11. Save best model weights
12. End

Based on Algorithm 1, the customized deep learning approach has been applied for feature extraction and regression. The pretrained backbone provided stable representations, while the added regression head is trained with MSE loss to predict MOS scores. This strategy was effective for transfer learning and improved the accuracy of contrast image quality assessment.

For further evaluation, we employed a Siamese network alongside pre-trained models to assess image contrast. Figure 2 illustrated a Siamese neural network using model-pertained backbones to compare two input images and predict their MOS difference.

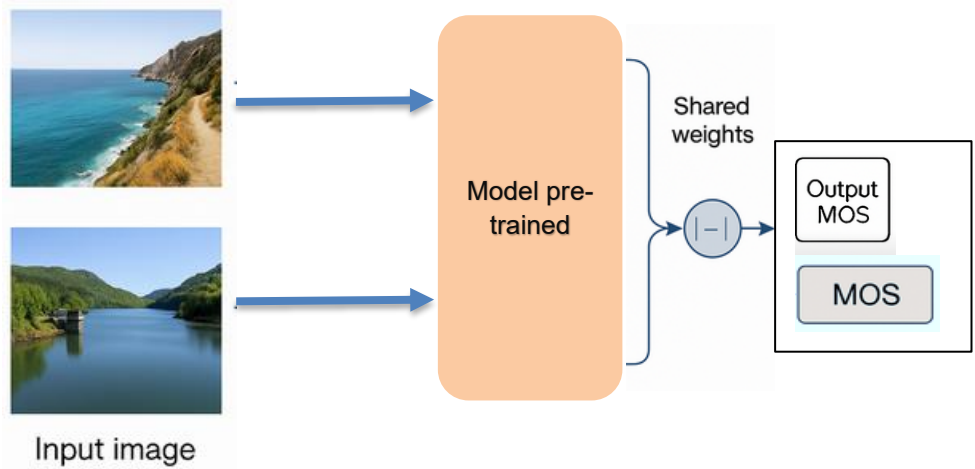

**Fig. 2.** Model Architecture with Siamese network.

*3.1. Data description*

The experiments were conducted on the Contrast Changed Image Database 2014 (CCID2014) [14] and Camera Image Database 2013 (CID2013) [43] . The CCID2014 dataset, as employed in our study, originally comprises 670 images; this total includes 15 undistorted original images and 655 contrasted versions. The contrasted images, each associated with a Mean Opinion Score (MOS), served as the ground truth for our regression task. We decided to exclude the 15 original reference images, also present in the full CCID2014 collection, from our experiments to focus on no-reference assessment. Before feeding the images to the network, the associated MOS values were normalized using a Z-score transformation. For this, the mean (μMOS) and standard deviation (σMOS) of all MOS values were calculated and then used to normalize each score:

$$Z_i = (MOS_i - \mu\_MOS) / \sigma\_MOS \qquad (1)$$

The predicted normalized scores are subsequently demoralized back to the original MOS scale for the final evaluation and interpretation of the performance.

The CID2013 dataset comprised 400 consumer camera images captured using 79 different devices across six distinct image sets. Each image was rated by up to 188 observers using a Dynamic Reference Absolute Category Rating (DR-ACR) protocol, and aggregated Mean Opinion Scores (MOS) serve as the ground truth for no-reference quality estimation. Designed to reflect real-world distortions like contrast variation and compression, making it both comprehensive and robust for assessing NR-IQA models [43].

Figure 3 showed Visual examples from the CCID2014 dataset. The top row displayed three original high-quality natural images selected from the dataset. The bottom row (d, e, f) presented corresponding versions of these images after contrast degradation using gamma correction with increasing distortion levels. These examples illustrated the dataset's structured approach to simulating contrast-related distortions while maintaining controlled variations, which was essential for training and evaluating contrast quality assessment models.

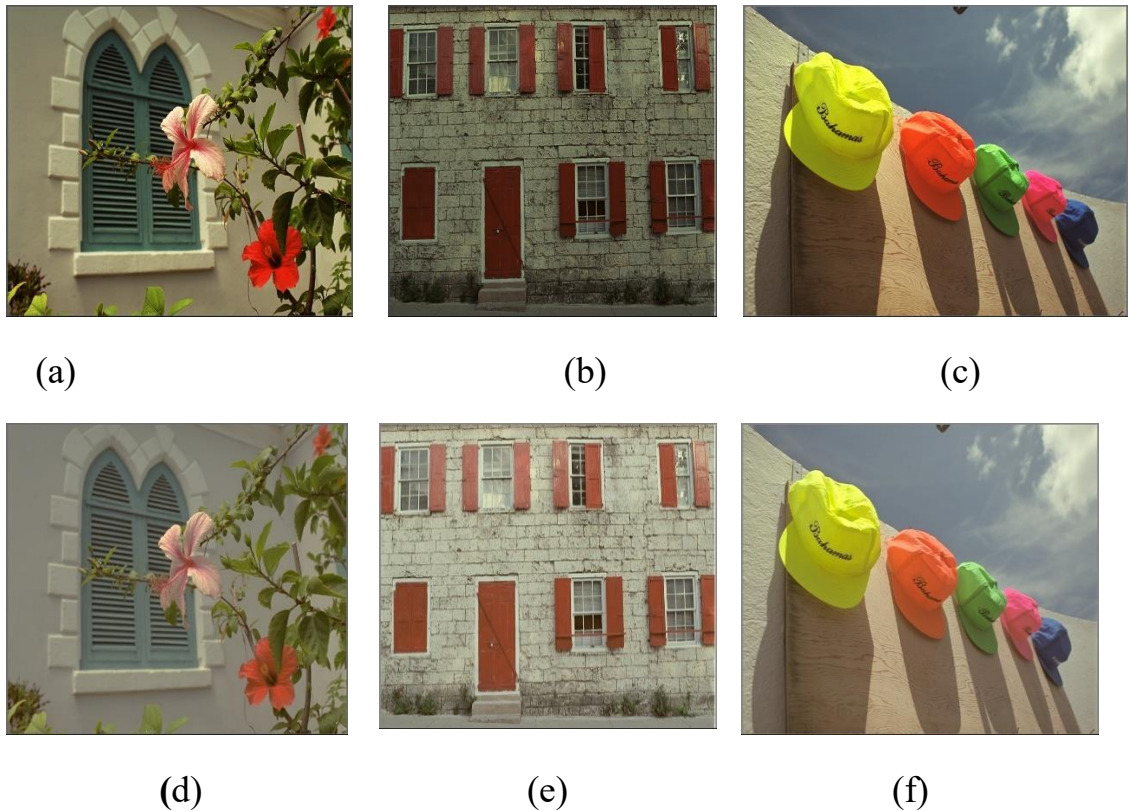

**Fig. 3.** The comparison of the original image (a, b, c) and the contrast-distorted images (d, e, f).

*3.2. Data Preprocessing*

The preprocessing stage played a vital role in preparing the dataset, particularly when images varied in size, contained noise, or were captured under inconsistent lighting conditions. Neglecting this stage may led to degraded image quality and reduced model performance. Preprocessing ensures that the dataset is standardized and optimized for effective training in convolutional neural networks (CNNs).

The first step in preprocessing was image resizing. Since the original dataset included images with different resolutions, such as 1920×1080, 640×480, and 720×120, the images were standardized by resizing them to 150×150 pixels. Although this resolution was relatively small, it reduced computational costs significantly while still maintaining essential features for training. As a result, the overall computational burden on the CNN was minimized.

Following resizing, normalization was applied to the dataset. In this process, pixel values were scaled into the range [0,1] through min–max scaling (rescale=1. 255). This normalization step stabilized the learning process, reduced sensitivity to varying light intensities across the dataset, and enhanced9the accuracy of the neural network by lowering the error rate during training.

Finally, to further improved the robustness of the model, online data augmentation was incorporated. Unlike offline augmentation methods that expanded the dataset beforehand, online augmentation applied transformations dynamically during training. In our preprocessing pipeline, we applied random rotations (up to 20 degrees), horizontal flips, zooming up to 20%, and horizontal and vertical shifts of 20%. These transformations ensured that every time an image is fed into the model, it may appear differently from previous iterations. This continuous variation prevented overfitting and enabled the model to generalize effectively to unseen data. By exposing the network to diverse orientations, scales, and positions of the same image, the preprocessing stage enhanced model robustness and improved performance on complex image quality assessment tasks.

The images underwent a comprehensive transformation pipeline designed not only to prepare them in the appropriate format for neural network input but also to artificially expand the diversity of the training set, thereby improving the model's ability to generalize to unseen data. The transformation sequence began with a random horizontal flip, applied with a 50% probability, to encourage the model to learn features invariant to left-right orientation. Next, each image was randomly rotated by an angle uniformly sampled from −10 to +10 degrees, enhancing robustness to small rotational changes.

To further increase the model's resilience against variations in lighting and color conditions, random color jitter was applied with adjustments to brightness (±20%), contrast (±20%), saturation (±20%), and hue (±0.1). Subsequently, all images were resized to a fixed dimension of 224×224 pixels to match the input size expected by the neural network. Finally, images were converted into PyTorch tensors and normalized using the mean [0.485, 0.456, 0.406] and standard deviation [0.229, 0.224, 0.225] values, consistent with ImageNet pre-trained model standards.

### 3.3. Proposed EfficientNet-B0 architecture

The architecture explored in this study is based on transfer learning with three widely used Convolutional Neural Networks (CNNs): EfficientNet-B0, ResNet50, and MobileNetV2. All three models were initialized with pre-trained weights from ImageNet and then fine-tuned for the task of image contrast quality assessment. After extensive evaluation, the EfficientNet-B0-based model exhibited the highest performance and therefore serves as the main backbone in our framework. Nevertheless, the comparative results with ResNet50 and MobileNetV2 provide valuable insights into the strengths and limitations of different architectures for this task.

The CNNs form the backbone of modern computer vision, designed with convolutional, pooling, and fully connected layers that progressively extract hierarchical features from input images. In transfer learning, pre-trained CNNs are repurposed for new tasks. This approach allows the network to retain low-level features (e.g., edges, textures) learned from large-scale datasets such as ImageNet, while higher-level layers are fine-tuned to adapt to the target domain. This greatly reduces the need for large labeled datasets and prevents overfitting.

In our architecture, all three backbones, EfficientNet-B0, ResNet50, and MobileNetV2, were incorporated in an end-to-end pipeline with a shared task-specific classification head. This design ensures that comparisons were fair, as the only differences lie in the backbone feature extractors. Among the tested models, EfficientNet-B0 provided superior performance. The key to its success lied in two innovations: compound scaling and MBConv blocks.

Unlike traditional scaling methods that arbitrarily increased depth, width, or input resolution, EfficientNet introduced a compound scaling method that jointly scaled all three dimensions with a single global coefficient $\phi$:

$$d = \alpha^\phi, \quad w = \beta^\phi, \quad r = \gamma^\phi \qquad (2)$$

subject to the constraint:

$$\alpha \cdot \beta^2 \cdot \gamma^2 \approx 2, \quad \alpha, \beta, \gamma \geq 1 \qquad (3)$$

Here:
- d was the network depth (number of layers).
- w was the network width (number of channels per layer).
- r was the input resolution.

This balanced approach ensured that increasing network size improved accuracy without excessively increasing computational cost. For EfficientNet-B0, the grid-searched coefficients were set to $\alpha = 1.2$, $\beta = 1.1$, and $\gamma = 1.15$, which results in an optimal trade-off between accuracy and efficiency compared to both shallower networks (e.g., EfficientNet-B0) and larger variants (e.g., B5–B7).

*3.4. ResNet50*

Resnet50 is another backbone evaluated in our study. ResNet's innovation lied in the concept of residual connections, which directly added the input of a block to its output, allowing the network to learn residual mappings instead of full transformations. This alleviated the vanishing gradient problem and enabled very deep networks to converge. The ResNet50 backbone comprised 50 layers, organized into convolutional blocks and identity blocks. Although ResNet50 was more computationally expensive than MobileNetV2, it provided robust performance for large-scale visual tasks. In our experiments, EfficientNet-B0 achieved reasonable performance and required lower training times and less memory than ResNet50.

*3.5. MobileNetV2*

MobileNetV2 was designed for mobile and embedded applications, where computational efficiency was critical. It used depth-wise separable convolutions, reducing the number of parameters and FLOPs compared to standard convolutions. Similar to EfficientNet-B0, it also employed inverted residual blocks (a lighter form of MBConv), but without the compound scaling mechanism. In our framework, MobileNetV2 proved to be the fastest model in terms of training and inference, but lagged behind both ResNet50 and EfficientNet-B0 in classification accuracy. This trade-off highlighted its strength in speed-sensitive scenarios, but for quality-sensitive tasks like image contrast assessment, the more balanced EfficientNet-B0 was more suitable.

On top of the backbone networks, a lightweight yet discriminative task-specific head was designed to adapt the extracted features to the image contrast quality classification task. First, a GlobalAveragePooling2D layer was employed to transform the spatial feature maps into compact descriptors, ensuring size invariance and reducing the number of trainable parameters. This was followed by one or more fully connected dense layers activated by ReLU, which captured high-level nonlinear interactions among features. To improve generalization and reduce the risk of overfitting, a dropout layer with a rate of 0.5 was integrated. Finally, a fully connected output layer with a single neuron and sigmoid activation function was appended to provide probabilistic predictions in the range [0,1]. This unified head design was consistently applied across EfficientNet-B0, ResNet50, and MobileNetV2, ensuring that performance differences were attributable to the backbones rather than the classification layers.

The training setup was carefully optimized to achieve fast convergence and robust generalization across all three backbones. The networks were trained in an end-to-end fashion using the Binary Cross-Entropy (BCE) loss function, which is well-suited for binary classification tasks such as high-versus low-quality contrast assessment. For optimization, the Adam optimizer was selected with an initial learning rate of $1\times10^{-4}$ times, providing adaptive moment estimation and stable updates. To prevent stagnation, a ReduceLROnPlateau scheduler was applied, which adaptively decreased the learning rate when validation accuracy plateaued. In addition, weight decay $1\times10^{-5}$ and dropout regularization were used to avoid overfitting, particularly important given the limited dataset size. Training was conducted with a batch size of 32, while ReLU activations were used in all hidden layers and sigmoid in the output layer. This configuration provided a balanced trade-off between computational efficiency and predictive accuracy, enabling the EfficientNet-B0 backbone to demonstrate its superior performance compared to ResNet50 and MobileNetV2.

### 3.3.1. Siamese Network

We have evaluated other models for image contrast assessment that are designed based on Siamese. These models employed a Siamese neural network architecture with EfficientNet-B0, ResNet-18, and MobileNetV2 as the shared backbone. Two distorted images were input into identical subnetworks of ImageNet. These subnetworks extracted 1280-dimensional feature vectors for each image. The absolute difference between these feature vectors captured perceptual dissimilarity related to image quality. This difference is passed through a regression head consisting of fully connected layers (512 → 256 → 1) with ReLU activations and dropout, outputting the predicted MOS. The model is fine-tuned end-to-end using MSE loss, enabling it to learn perceptual quality estimation tailored to contrast-distorted images in the CCID2014 and CID2013 datasets. Figure 4 illustrated a Block diagram of a Siamese network architecture for full-reference contrast quality assessment, where paired CNN branches extracted features from reference and distorted images to predict perceived contrast quality.

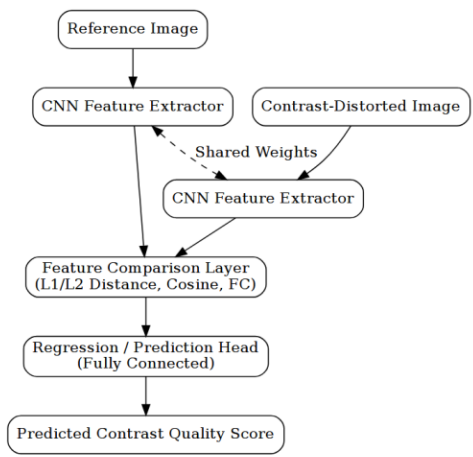

**Fig 4.** Siamese network architecture.

*3.4. Finetuning hyperparameters*

To ensure reproducibility and provide insight into the training process, the hyperparameters used for training the MOS prediction model with EfficientNet-B0, ResNet18, and MobileNetV2 are summarized in Table 2. These parameters were chosen based on common deep learning practices and were fine-tuned through preliminary experiments to achieve optimal performance on the CCID2014 and CID2013 datasets.

**Table 2.** Hyperparameters used in Model Training.

| Hyperparameter | Value | Description |
|---|---|---|
| Optimizer | Adam | Adaptive moment estimation optimizer used for gradient descent. |
| Learning Rate | $1 \times 10^{-4}$ | Initial learning rate for the Adam optimizer. |
| Loss Function | Mean Squared Error (MSE) | Measures the squared difference between predicted and actual MOS values. |
| Epochs | 50 | Total number of training iterations over the dataset. |
| Batch Size | 32 | Number of samples processed before the model is updated. |
| Fully Connected Layers | $1280 \rightarrow 512 \rightarrow 256 \rightarrow 1$ | Layer sizes used in the regression head after EfficientNet-B0. |
| Backbone Model | EfficientNet-B0 (frozen) ResNet-18 (frozen) MobileNetV2 (frozen) | Pretrained CNN used for feature extraction with an output size of 1280. |
| Input Image Size | $224 \times 224$ | Final image size after resizing. |
| Data Augmentation | RandomHorizontalFlip, Rotation (±10°), ColorJitter, Resize, Normalize | Augmentations are used to increase training diversity. |
| Normalization Mean | [0.485, 0.456, 0.406] | Mean values for normalization across RGB channels. |
| Normalization Std | [0.229, 0.224, 0.225] | Standard deviation for normalization across RGB channels. |
| Weight Decay | 1e-5 | Regularization factor used in the optimizer to prevent overfitting. |
| Scheduler | ReduceLROnPlateau | Learning rate scheduler that reduces LR when validation loss plateaus. |
| Device | CUDA | Training executed on GPU using CUDA backend. |
| Evaluation Metric | Tolerance Accuracy (±0.5), Pearson & Spearman Correlation | Used to assess prediction accuracy and correlation with ground truth. |

## 4. Results

In the results, we compared different CNN backbones and Siamese networks on the CID2013 and CCID2014 datasets for contrast quality prediction. The evaluation showed that EfficientNet-B0 achieved the strongest alignment with subjective MOS, while conventional Siamese setups delivered only moderate correlations, highlighting the need for more advanced designs.

*4.1. Experimental Setup*

The CCID2014 and CID2013 datasets, consisting of contrast-distorted images and corresponding MOS values, were split into training (80%) and validation (20%) sets. Without augmentation, images were preprocessed using standard normalization (mean: [0.485, 0.456, 0.406], std: [0.229, 0.224, 0.225]). With augmentation, transformations such as random horizontal flipping, rotation, and color jitter were applied to increase data diversity. The model was trained for 50 epochs using the Mean Squared Error (MSE) loss function and the Adam optimizer (learning rate: $1 \times 10^{-4}$, weight decay: 1e-5). Evaluation accuracy was defined as the percentage of predictions within ±0.5 of the actual MOS, providing a perceptually meaningful measure of prediction reliability.

*4.2. Comparison of the models on the CCID 2014 dataset*

In this section, we present a comparative analysis of different CNN architectures and Siamese networks for contrast-distorted image quality prediction on the CCID2014 dataset. The evaluation covered ResNet-18, EfficientNet-B0, and MobileNetV2 in both custom pretrained and Siamese configurations, comparing their PLCC and SRCC, correlation with MOS, and generalization performance to identify the most effective approach for perceptual contrast assessment.

*4.2.1. Results with custom pre-trained model*

Table 2 presented the performance evaluation of three convolutional neural network models—ResNet-18, EfficientNet-B0, and MobileNetV2 developed for image contrast assessment using predicted Mean Opinion Scores (MOS). Among the models, ResNet-18 demonstrated superior validation accuracy (87.64%) and maintained a strong correlation with MOS values (Pearson: 0.8669, Spearman: 0.8443), indicating its effectiveness in capturing perceptual contrast features. Although EfficientNet-B0 shows lower validation accuracy (66.41%), it achieved the highest correlation coefficients (Pearson: 0.9286, Spearman: 0.9178), suggesting better alignment with human visual perception. In contrast, MobileNetV2, while having a lower validation accuracy (60.31%), offers comparable correlation values, making it a lightweight yet competitive alternative for contrast prediction tasks where computational efficiency was critical.

Figure 5 illustrates the performance comparison of the three models used:

EfficientNet-B0 demonstrated the most stable and balanced training behavior. While training loss steadily decreased and training accuracy exceeded 95%, the validation accuracy remained around 75% with minimal fluctuation, indicating good generalization without overfitting.

ResNet-18 achieved the highest training accuracy (~99%) and lowest training loss, but exhibited a clear gap between training and validation curves. The validation accuracy plateaued around 88%, suggesting slight overfitting despite strong overall performance.

MobileNetV2, although lightweight, showed the lowest validation accuracy (~65%) and exhibited the largest gap between training and validation performance, indicating significant overfitting and limited generalization capability.

In terms of the MOS scatter plots:

**EfficientNet-B0** presented the tightest clustering of predicted vs. real MOS values along the diagonal trend line, showing high linear correlation and prediction consistency.

**ResNet-18** also showed strong MOS prediction performance, but with slightly more dispersion around the trend line.

**MobileNetV2** revealed more spread in predictions, particularly in the mid-range MOS values, implying lower regression precision.

In summary, EfficientNet-B0 offered the best trade-off between model complexity and generalization for the image contrast assessment task.

Based on the actual versus predicted MOS values presented in Table 3, the proposed EfficientNet-B0 model demonstrated the highest consistency with the ground-truth MOS compared to the other models. While all models show reasonable alignment, EfficientNet-B0 exhibited superior ability in predicting human-perceived image quality on the CCID2014 dataset, confirming its effectiveness for continuous contrast quality estimation.

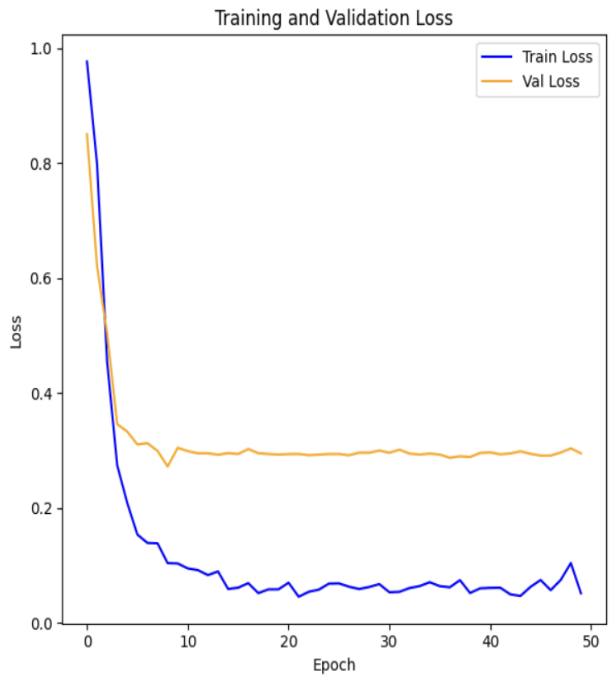
(a) EfficientNet-B0: Loss

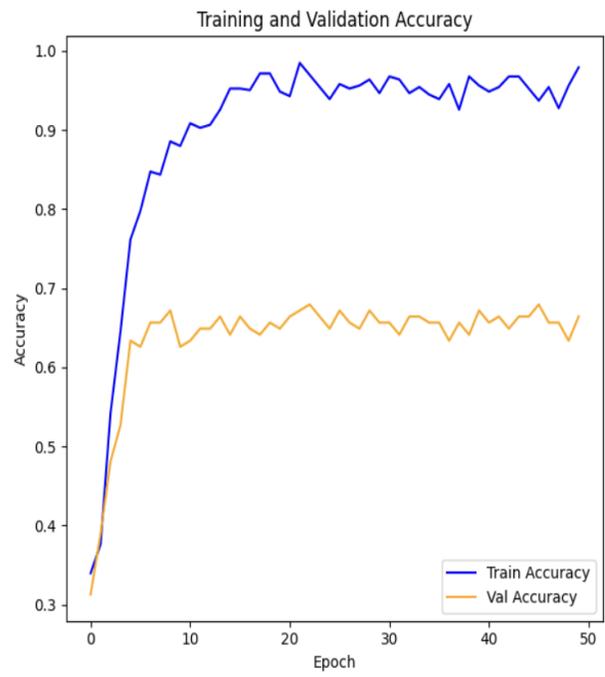
(b) EfficientNet-B0: Acc

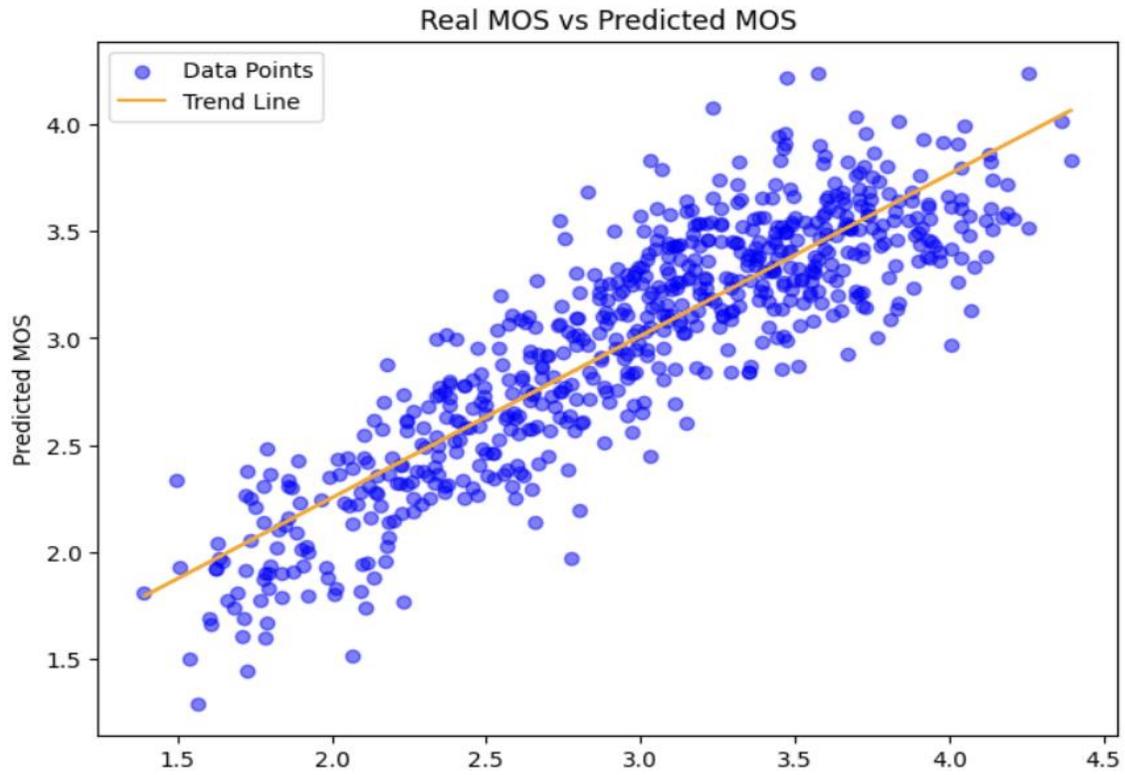

(d) EfficientNet-B0: MOS

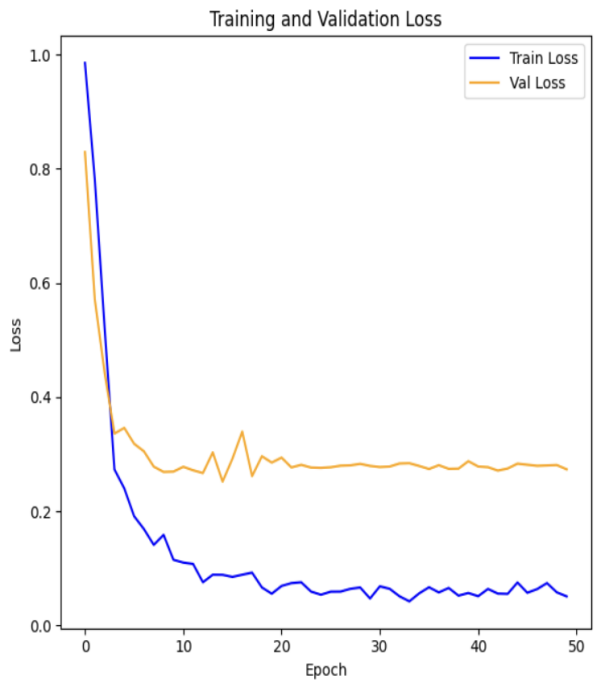

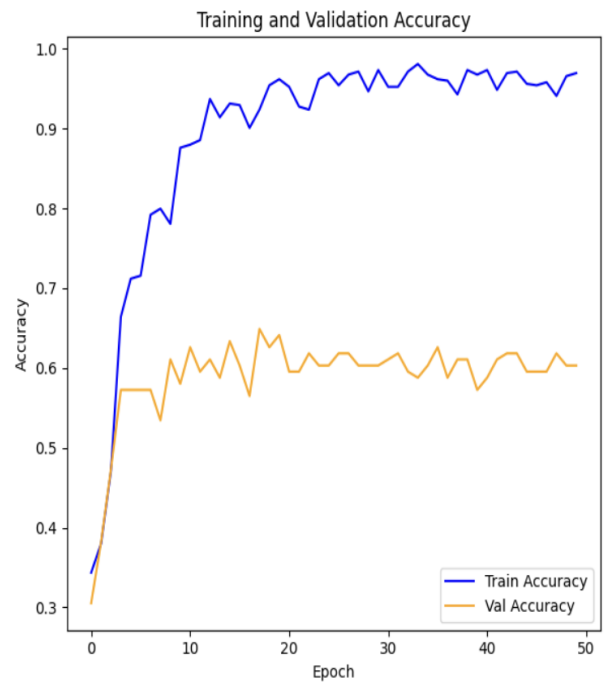

(e) MobileNet: Loss  (f) MobileNet: Acc

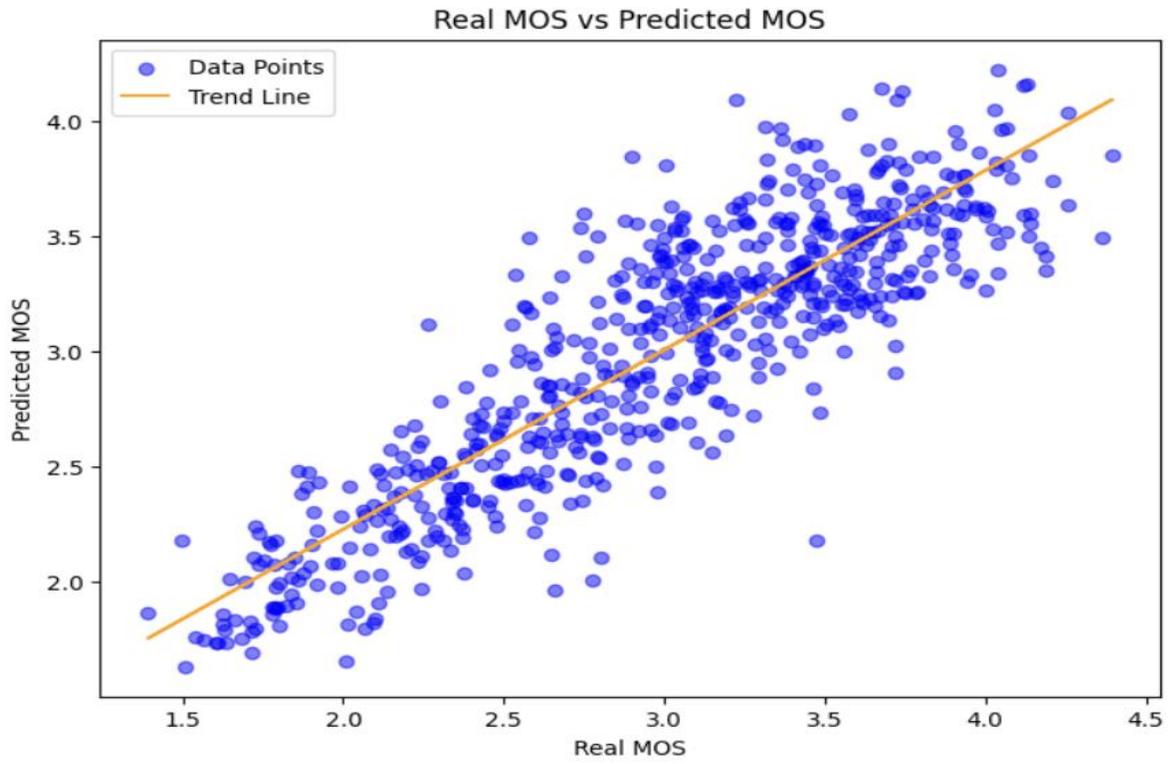

(g) MobileNet: MOS

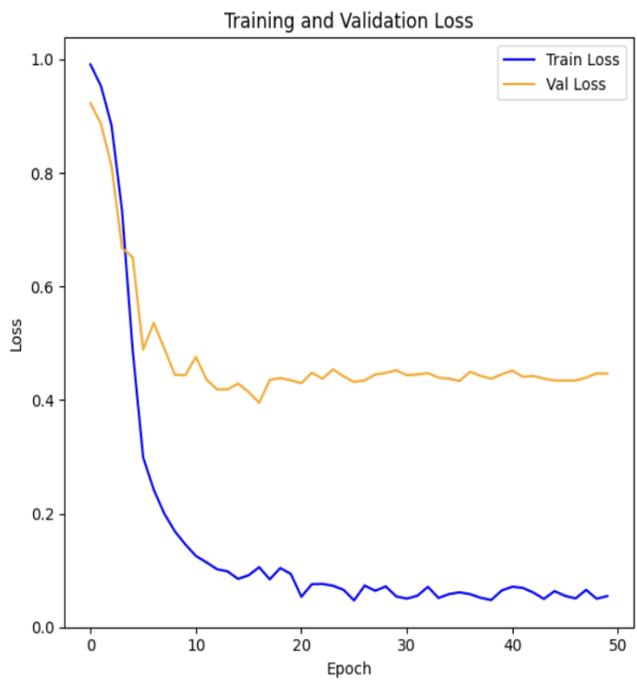

(h) ResNet-18: Loss

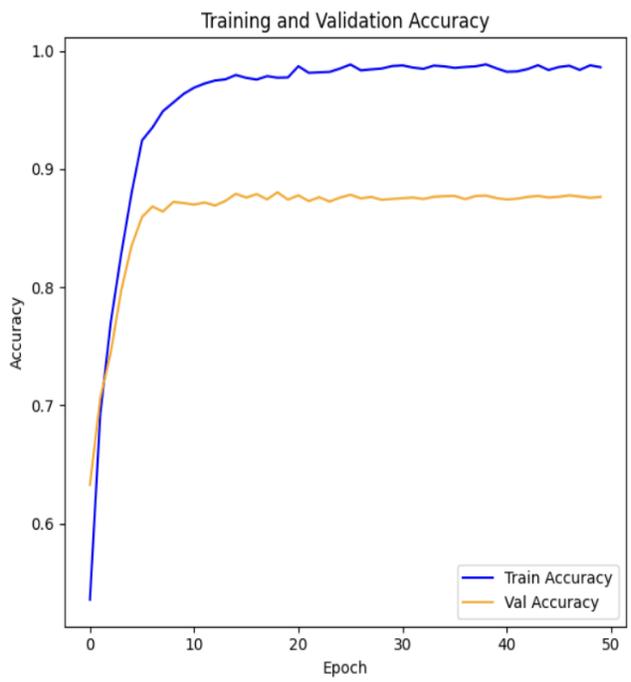

(i) ResNet-18: Acc

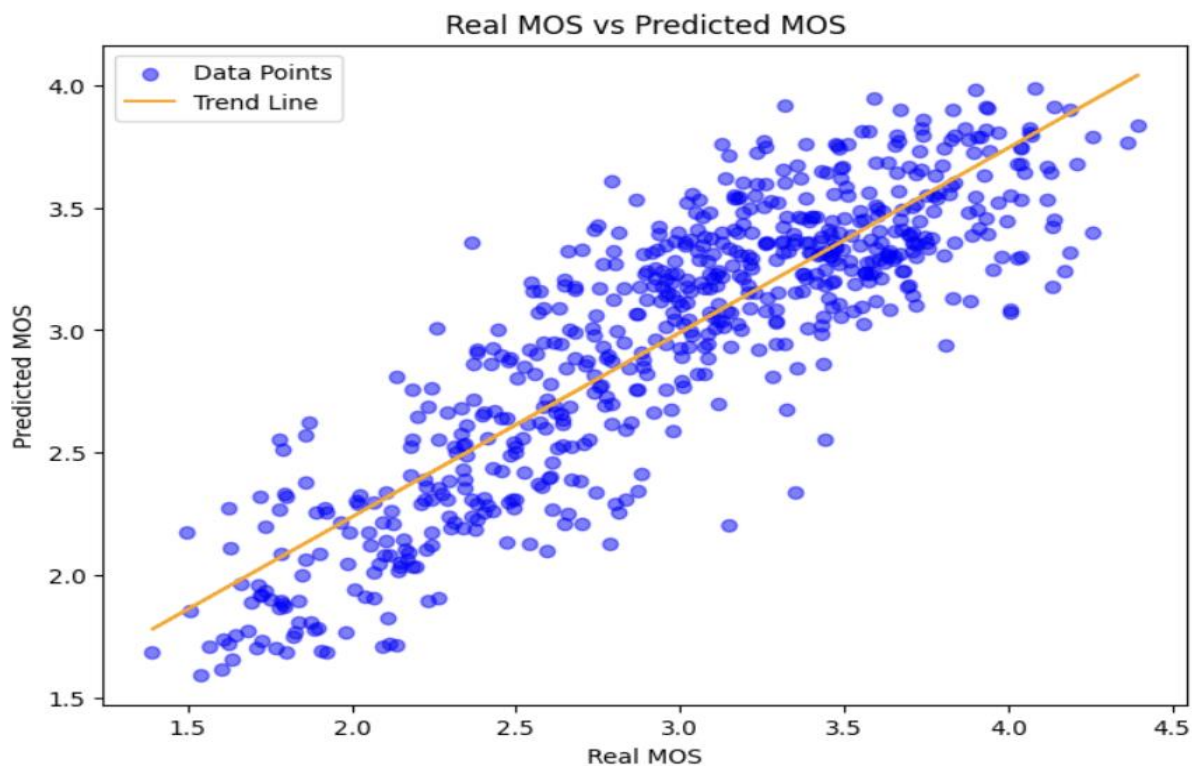

(j) ResNet-18: MOS

Fig.5 Performance comparison of ResNet-18: (h), (i), (j), MobileNetV2: (e), (f), (g), and EfficientNet-B0: (a), (b), (d)

Based on the results in Table 4, where the goal was continuous quality prediction, EfficientNet-B0 and MobileNetV2 outperform ResNet-18 in terms of Pearson and Spearman correlation coefficients, indicating better alignment with human perceptual scores. Therefore, EfficientNet-B0 was the most appropriate model for predicting contrast-distorted image quality in a way that aligns with human judgment.

Table 3: Performance Metrics for Custom Pretrained Model

| Model | PLCC | Spearman Corr |
|---|---|---|
| ResNet-18 | 0.8669 | 0.8443 |
| EfficientNet-B0 | 0.9286 | 0.9178 |
| MobileNetV2 | 0.8746 | 0.8545 |

Table 4: Actual vs Predicted MOS for ResNet-18, EfficientNet-B0, and MobileNetV2

| Index | Actual MOS (ResNet-18) | Predicted MOS (ResNet-18) | Actual MOS (EfficientNet-B0) | Predicted MOS (EfficientNet-B0) | Actual MOS (MobileNetV2) | Predicted MOS (MobileNetV2) |
|---|---|---|---|---|---|---|
| 0 | 3.1577 | 3.2848 | 3.1577 | 3.2848 | 3.1577 | 3.2848 |
| 1 | 2.7415 | 2.7520 | 2.7415 | 2.7520 | 2.7415 | 2.7520 |
| 2 | 2.2586 | 2.3284 | 2.2586 | 2.3284 | 2.2586 | 2.3284 |
| 3 | 1.6939 | 1.8089 | 1.6939 | 1.8089 | 1.6939 | 1.8089 |
| 4 | 2.3786 | 2.7838 | 2.3786 | 2.7838 | 2.3786 | 2.7838 |
| 5 | 2.7239 | 2.6809 | 2.7239 | 2.6809 | 2.7239 | 2.6809 |
| 6 | 2.9562 | 2.8135 | 2.9569 | 2.8135 | 2.9562 | 2.8135 |
| 7 | 3.1271 | 3.0594 | 3.1271 | 3.0594 | 3.1271 | 3.0594 |
| 8 | 1.7768 | 2.3111 | 1.7768 | 2.3111 | 1.7768 | 2.3111 |
| 9 | 2.0197 | 2.4335 | 2.0197 | 2.4335 | 2.0197 | 2.4335 |
| 10 | 2.4934 | 2.7320 | 2.4934 | 2.7320 | 2.4934 | 2.7320 |
| 11 | 2.9255 | 3.3284 | 2.9255 | 3.3284 | 2.9255 | 3.3284 |
| 12 | 3.4678 | 3.4956 | 3.4678 | 3.4956 | 3.4678 | 3.4956 |
| 13 | 3.3814 | 3.5618 | 3.3814 | 3.5618 | 3.3814 | 3.5618 |
| 14 | 3.3133 | 3.7219 | 3.3133 | 3.7219 | 3.3133 | 3.7219 |
| 15 | 3.2113 | 3.2797 | 3.2113 | 3.2797 | 3.2113 | 3.2797 |
| 16 | 2.9272 | 3.2044 | 2.9272 | 3.2044 | 2.9272 | 3.2044 |
| 17 | 2.8986 | 3.0960 | 2.8986 | 3.0960 | 2.8986 | 3.0960 |
| 18 | 2.5918 | 2.6238 | 2.5918 | 2.6238 | 2.5918 | 2.6238 |
| 19 | 2.2338 | 2.7379 | 2.2338 | 2.7379 | 2.2338 | 2.7379 |
| 20 | 3.3079 | 3.4435 | 3.3079 | 3.4435 | 3.3079 | 3.4435 |
| 21 | 3.0713 | 3.1066 | 3.0713 | 3.1066 | 3.0713 | 3.1066 |
| 22 | 2.6614 | 2.5867 | 2.6614 | 2.5867 | 2.6614 | 2.5867 |
| 23 | 2.1809 | 2.1392 | 2.1809 | 2.1392 | 2.1809 | 2.1392 |
| 24 | 2.0122 | 1.8325 | 2.0122 | 1.8325 | 2.0122 | 1.8325 |
| 25 | 1.7101 | 1.6081 | 1.7101 | 1.6083 | 1.7101 | 1.6081 |
| 26 | 3.3484 | 3.3911 | 3.3484 | 3.3911 | 3.3484 | 3.3911 |
| 27 | 3.6542 | 3.6849 | 3.6542 | 3.6849 | 3.6542 | 3.6849 |
| 28 | 4.1364 | 3.8226 | 4.1364 | 3.8226 | 4.1364 | 3.8226 |
| 29 | 3.4700 | 3.4917 | 3.4700 | 3.4917 | 3.4700 | 3.4917 |
| 130 | 3.5621 | 3.3431 | 3.5621 | 3.3431 | 3.5621 | 3.3431 |
| 31 | 4.1342 | 3.6061 | 4.1342 | 3.6061 | 4.1342 | 3.6061 |
| 32 | 3.9392 | 3.5597 | 3.9392 | 3.5597 | 3.9392 | 3.5597 |
| 33 | 3.7100 | 3.6500 | 3.7100 | 3.6500 | 3.7100 | 3.6500 |
| 34 | 3.4721 | 4.2139 | 3.4721 | 4.2139 | 3.4721 | 4.2139 |
| 35 | 3.6714 | 3.8215 | 3.6714 | 3.8215 | 3.6714 | 3.8215 |
| 36 | 4.2585 | 3.5162 | 4.2585 | 3.5162 | 4.2585 | 3.5162 |
| 37 | 3.4635 | 3.8844 | 3.4635 | 3.8844 | 3.4635 | 3.8844 |

*4.2.2. Results with Siamese network*

The evaluated Siamese architectures EfficientNet-B0, MobileNetV2, and ResNet18 are trained in a pairwise regression setting to predict MOS under contrast distortion conditions. Performance was quantified using the PLCC, which measures prediction MOS linear agreement after nonlinear regression mapping, and the SRCC, which evaluated the monotonicity of the ranking produced by the model. Table 5 presented the Pearson and Spearman correlation coefficients for the evaluated Siamese backbones on the CCID2014 contrast distortion dataset.

From a technical standpoint, these correlation levels fall within the low-to-moderate predictive regime. In the context of FR IQA literature, state-of-the-art contrast distortion predictors typically achieved PLCC and SRCC values $\geq 0.8$ on CCID2014 or comparable datasets, following monotonic mapping. The observed gap suggested that the current Siamese training configuration was not extracting or leveraging sufficiently discriminative features for contrast degradation.

**Table 5.** Performance Metrics for Siamese Network.

| Model | PLCC | SRCC |
|---|---|---|
| ResNet-18 | 0.4475 | 0.4397 |
| EfficientNet-B0 | 0.4908 | 0.5078 |
| MobileNetV2 | 0.4530 | 0.4493 |

The moderate result likely stems from architectural, training, and dataset limitations, making the current Siamese models unsuitable for reliable contrast distortion assessment without further optimization in backbone design, loss functions, data diversity, and score calibration.

## 4.3 Comparison of the models on the CID 2013 dataset

This section presented a comparative analysis of ResNet-18, EfficientNet-B0, and MobileNetV2 on the CID2013 dataset using both custom pretrained and Siamese configurations for contrast-distorted image quality prediction. Results highlighted EfficientNet-B0 as the most perceptually aligned model in the custom pretrained setting, while Siamese networks, despite EfficientNet-B0's relative lead, showed only moderate correlation with subjective MOS, indicating the need for architectural and training enhancements.

*4.3.1. Results with custom pretrained model*

Figure 6 illustrated training and validation loss, accuracy, and Real vs. Predicted MOS for EfficientNet-B0, ResNet-18, and MobileNetV2 on the CID2013 dataset.

All three models showed effective learning behavior; however, EfficientNet-B0 demonstrated the strongest linear correlation between predicted and actual MOS, as seen in the third plot of each model. Despite ResNet-18 showing higher validation accuracy, EfficientNet-B0 achieved better perceptual alignment with human scores, making it a more reliable model for continuous image contrast quality estimation.

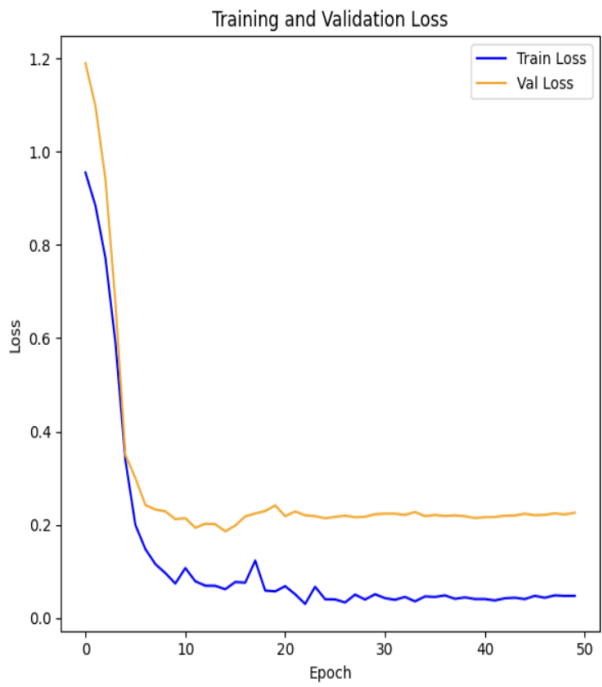

(a) EffNet: Los

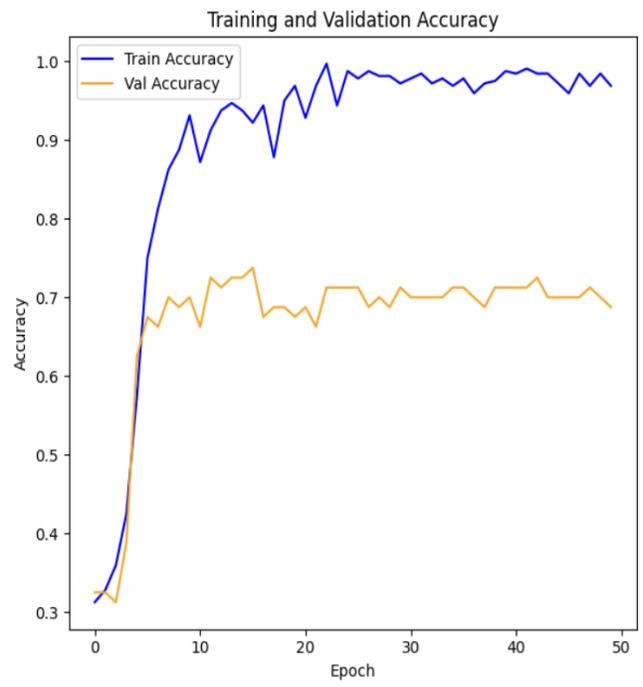

(b) EffNet: Acc

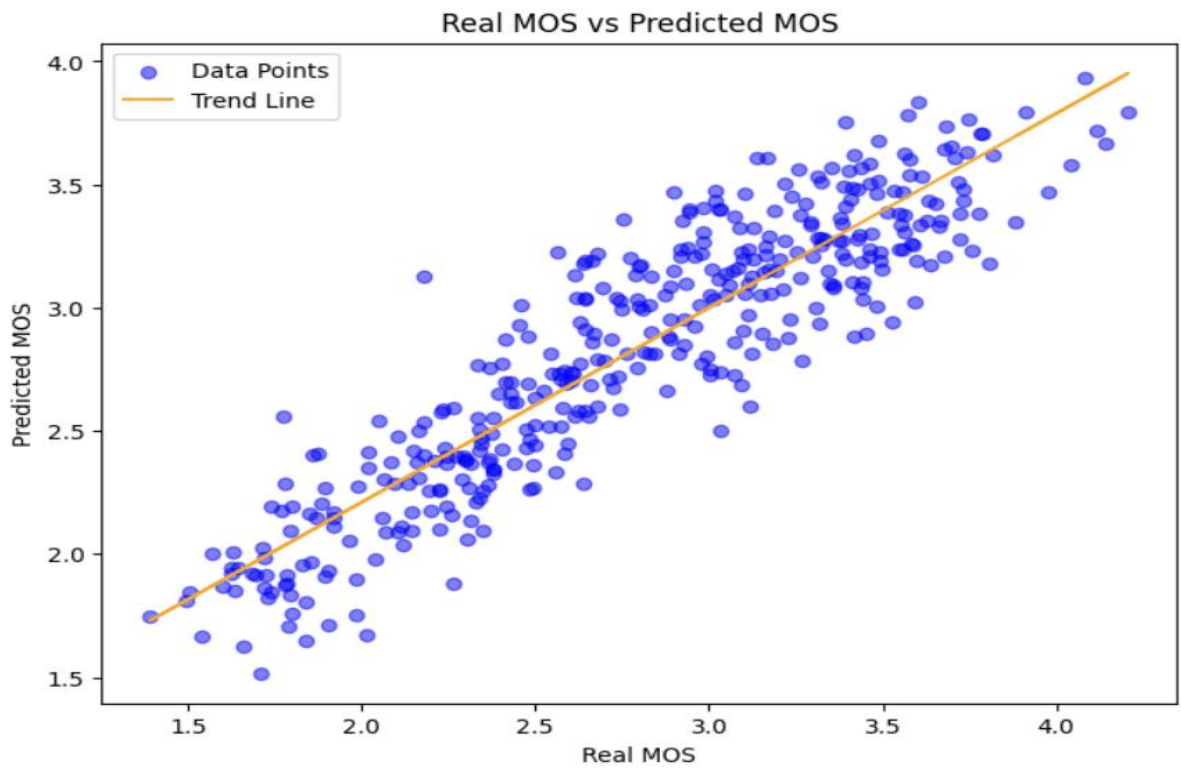

(c) EffNet: MOS

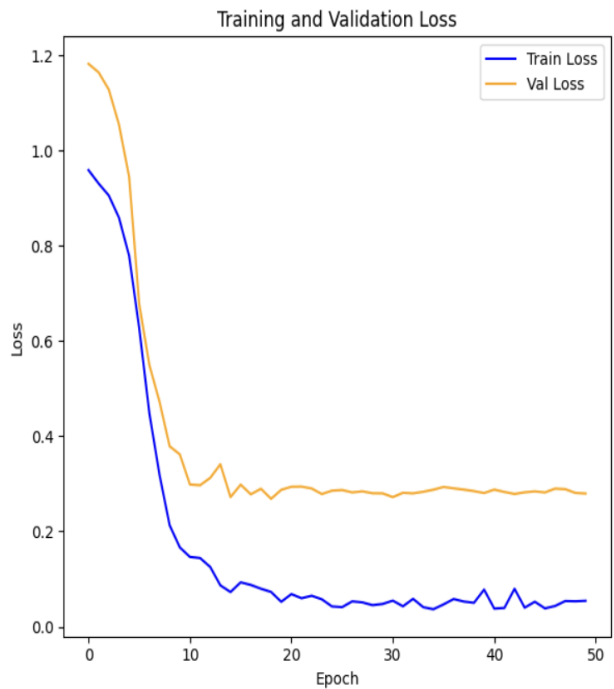

(d) ResNet-18: Loss

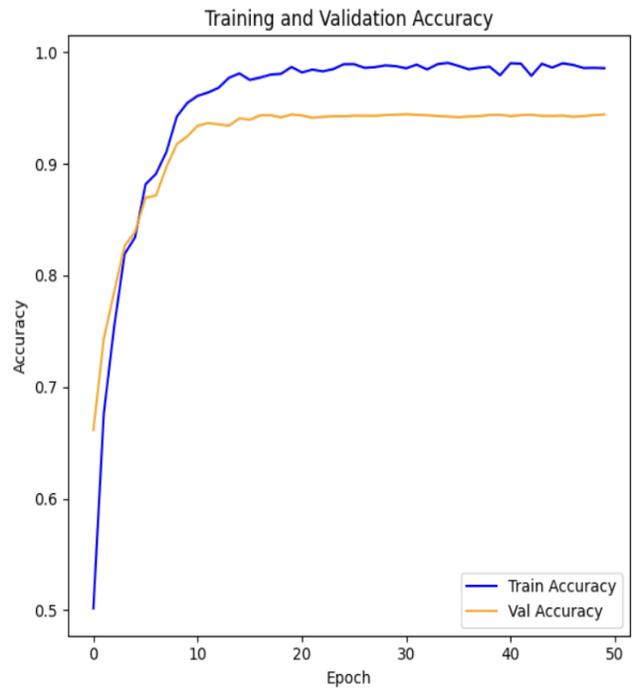

(e) ResNet-18: Acc

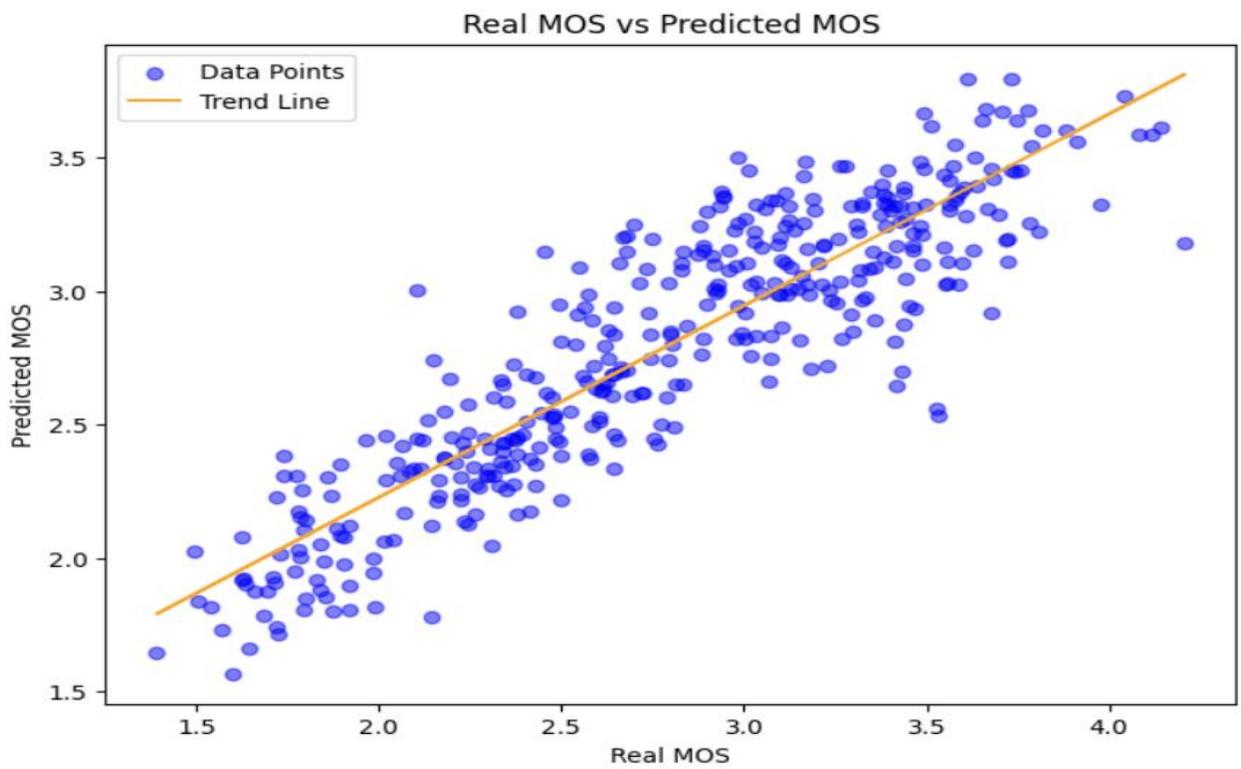

(f) ResNet-18: MOS

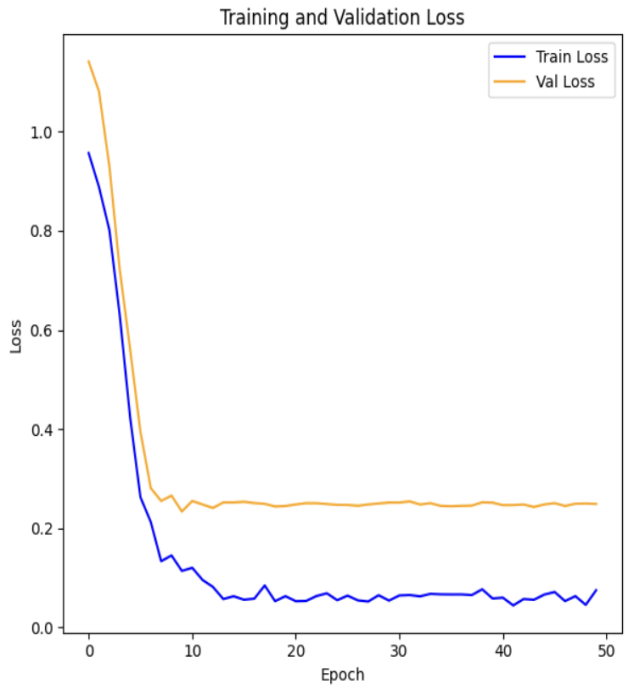
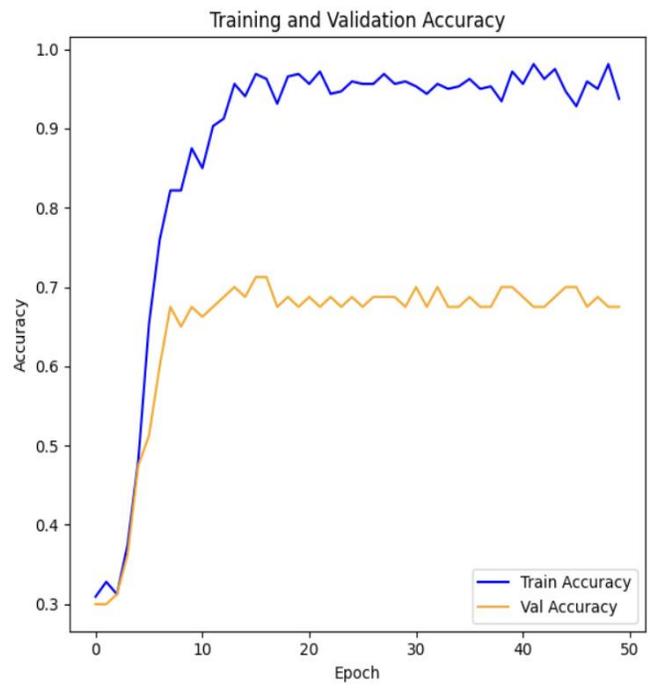

(g) MobileNet: Loss

(h) MobileNet: Acc

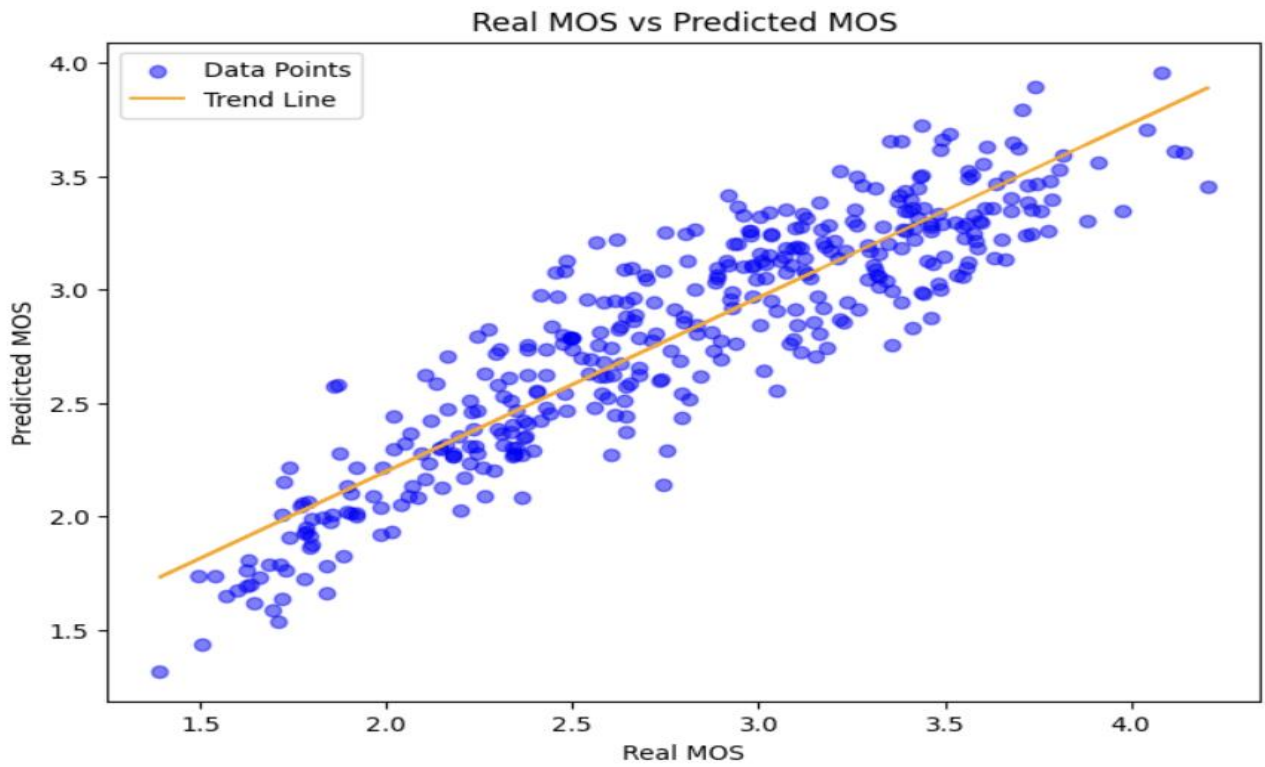

(i) MobileNet: MOS

23Fig. 6 Performance comparison of ResNet-18: (d), (e), (f), MobileNetV2: (g), (h), (i), and EfficientNet-B0: (a), (b), (c).

Table 6 provided a quantitative comparison of three deep learning architectures:

- **EfficientNet-B0** achieved the highest Pearson (0.9581) and Spearman (0.9369) correlations, which directly measure the model's ability to predict MOS values in a linear and monotonic manner, respectively. This was critical since image contrast assessment was inherently a perceptual, regression-based task.
- **MobileNetV2**, despite weak classification models, slightly outperforms ResNet18 in both Pearson (0.9055 vs. 0.8919) and Spearman (0.8969 vs. 0.8809) correlations, suggesting its regression predictions were relatively reliable, though it may underperform in classification tasks.
- **ResNet-18**, while excelling in accuracy, showed the lowest correlation values, indicating that its predictions were less aligned with human-perceived contrast scores despite high classification performance.

EfficientNet-B0 provided the best overall trade-off for image contrast assessment. It combined low training and validation loss with the strongest correlation to real MOS scores, indicating both robust feature learning and high perceptual alignment. ResNet-18, though highly accurate in classification, was prone to overfitting and exhibited weaker regression consistency. MobileNetV2, while efficient, was less suitable due to lower accuracy and limited learning depth, though its correlation models are still competitive. These findings validate the suitability of using a customized EfficientNet-B0 for contrast quality prediction, especially in scenarios where perceptual alignment with human judgments, MOS, is more critical than raw classification accuracy.

Table 6. Performance Metrics for custom pretrained Models.

| Model | PLCC | SRCC |
| --- | --- | --- |
| ResNet-18 | 0.8919 | 0.8809 |
| EfficientNet-B0 | 0.9581 | 0.9369 |
| MobileNetV2 | 0.9055 | 0.8969 |

Table 7 presented a detailed comparison between actual and predicted MOS values for ResNet18, EfficientNet-B0, and MobileNetV2 on the CID2013 dataset

Table 7. Actual vs Predicted MOS for ResNet-18, EfficientNet-B0 and MobileNetV2.

| Index | Actual MOS (ResNet-18) | Predicted MOS(ResNet-18) | Actual MOS (EfficientNet-B0) | Predicted MOS (EfficientNet-B0) | Actual MOS (MobileNetV2) | Predicted MOS (MobileNetV2) |
| --- | --- | --- | --- | --- | --- | --- |
| 0 | 3.1577 | 3.0583 | 3.1577 | 3.1449 | 3.1577 | 2.9731 |
| 1 | 2.7415 | 2.8374 | 2.7415 | 2.5886 | 2.7415 | 3.0829 |
| 2 | 2.2586 | 2.3435 | 2.2586 | 2.1575 | 2.2586 | 2.2162 |
| 3 | 1.6939 | 1.8770 | 1.6939 | 1.9180 | 1.6939 | 1.5895 |
| 4 | 2.3786 | 2.1624 | 2.3786 | 2.3417 | 2.3786 | 2.4134 |
| 5 | 2.7239 | 3.1535 | 2.7239 | 2.6732 | 2.7239 | 2.8047 |
| 6 | 2.9562 | 3.0100 | 2.9569 | 2.9228 | 2.9562 | 3.1053 |
| 7 | 3.1271 | 2.1725 | 3.1271 | 3.1970 | 3.1271 | 3.0745 |
| 8 | 1.7768 | 2.2955 | 1.7768 | 2.2843 | 1.7768 | 1.9242 |
| 9 | 2.0197 | 2.9494 | 2.0197 | 2.3481 | 2.0197 | 2.2995 |
| 10 | 2.4934 | 3.0039 | 2.4934 | 2.3619 | 2.4934 | 2.7903 |
| 11 | 2.9255 | 2.9323 | 2.9255 | 2.9504 | 2.9255 | 2.9894 |
| 12 | 3.4678 | 3.3283 | 3.4678 | 3.3000 | 3.4678 | 3.1177 |
| 13 | 3.3814 | 3.0441 | 3.3814 | 3.2720 | 3.3814 | 3.1847 |
| 14 | 3.3133 | 3.0254 | 3.3133 | 3.2854 | 3.3133 | 3.0892 |
| 15 | 3.2113 | 3.0265 | 3.2113 | 3.0718 | 3.2113 | 3.1405 |
| 16 | 2.9272 | 2.9528 | 2.9272 | 3.8498 | 2.9272 | 2.9177 |
| 17 | 2.8986 | 2.6345 | 2.8986 | 3.1525 | 2.8986 | 2.7778 |
| 18 | 2.5918 | 2.1389 | 2.5918 | 2.4467 | 2.5918 | 2.5267 |
| 19 | 2.2338 | 3.2529 | 2.2338 | 2.5868 | 2.2338 | 2.3869 |
| 20 | 3.3079 | 2.8353 | 3.3079 | 3.0007 | 3.3079 | 3.1088 |
| 21 | 3.0713 | 2.6991 | 3.0713 | 2.8618 | 3.0713 | 3.1823 |
| 22 | 2.6614 | 2.3807 | 2.6614 | 3.1886 | 2.6614 | 2.8664 |
| 23 | 2.1809 | 2.0614 | 2.1809 | 2.5347 | 2.1809 | 2.2667 |
| 24 | 2.0122 | 1.9277 | 2.0122 | 1.6691 | 2.0122 | 1.9353 |
| 25 | 1.7101 | 2.8917 | 1.7101 | 1.5167 | 1.7101 | 1.5395 |
| 26 | 3.3484 | 3.0861 | 3.3484 | 3.0827 | 3.3484 | 2.9940 |
| 27 | 3.6542 | 2.7007 | 3.6542 | 3.1501 | 3.6542 | 3.0419 |
| 28 | 4.1364 | 2.8909 | 4.1364 | 3.2921 | 4.1364 | 3.4984 |
| 29 | 3.4700 | 1.8008 | 3.4700 | 3.2216 | 3.4700 | 3.2414 |
| 30 | 3.5621 | 2.4399 | 3.5621 | 3.8706 | 3.5621 | 2.9776 |
| 31 | 4.1342 | 2.7035 | 4.1342 | 3.4051 | 4.1342 | 2.2784 |
| 32 | 3.9392 | 3.0786 | 3.9392 | 3.2669 | 3.9392 | 2.7914 |
| 33 | 3.7100 | 3.3218 | 3.7100 | 3.5974 | 3.7100 | 2.7879 |
| 34 | 3.4721 | 2.3361 | 3.4721 | 4.9359 | 3.4721 | 3.0672 |
| 35 | 3.6714 | 2.2697 | 3.6714 | 3.1044 | 3.6714 | 2.8327 |
| 36 | 4.2585 | 2.8898 | 4.2585 | 2.1107 | 4.2585 | 2.2318 |
| 37 | 3.4635 | 2.2865 | 3.4635 | 2.2093 | 3.4635 | 2.6129 |

### *4.3.2. Results with the Siamese network*

Siamese networks, as classically implemented, exhibited structural limitations for reliable contrast distortion assessment, and the demonstrated need for attention mechanisms and tailored loss functions in more advanced models underscores that conventional Siamese architectures and training objectives did not inherently align with perceptual ranking or MOS regression tasks, rendering them suboptimal in their vanilla form .Table 8 summarized the PLCC and SRCC values for three Siamese backbones, showing that although EfficientNet-B0 attained the highest correlations, all models demonstrated only moderate alignment with subjective MOS for contrast distortion assessment**.**

**Table 8.** Performance Metrics for Siamese Network

| Model | PLCC | SRCC |
|---|---|---|
| ResNet-18 | 0.4475 | 0.4397 |
| EfficientNet-B0 | 0.4908 | 0.5078 |
| MobileNetV2 | 0.4530 | 0.4493 |

### *4.4. Loss and Evaluation Metrics*

We employed Mean Squared Error (MSE) loss with other evaluation metrics model performance was quantified using two standard metrics, calculated by comparing the denormalized and clipped predicted MOS values against the original ground-truth MOS values from the validation set:

We review the performance of three standard criteria in non-reference quality assessment:

- PLCC was a statistical measure used to quantify the strength and direction of the linear relationship between two variables. In the field of Image Quality Assessment (IQA), it was commonly used to evaluate the similarity between the predicted quality scores generated by a model and the ground-truth scores (such as MOS or DMOS) obtained from human observers. The PLCC was defined by the following formula:

$$\text{PLCC} = [\Sigma(x_i - \bar{x})(y_i - \bar{y})] / [\text{sqrt}(\Sigma(x_i - \bar{x})^2) * \text{sqrt}(\Sigma(y_i - \bar{y})^2)] \qquad (5)$$

SRCC was widely used to compare the relative ranking of predicted quality scores from a model with the ground-truth rankings (e.g., MOS or DMOS) assigned by human observers, in the context of IQA. A higher SRCC value indicated better consistency in ranking between the predicted and subjective scores. The SRCC is defined by the following formula:

$$\text{SRCC} = 1 - [6 \times \Sigma d_i^2] / [N \times (N^2 - 1)] \qquad (6)$$

### 4.5. Comparison with other models

Table 9 presented a comprehensive comparison of the proposed EfficientNet-B0-based NR-IQA method against several RR and NR models on the CID2013 and CCID2014 datasets. The evaluation is conducted using two standard correlation metrics: PLCC and SRCC, which measure prediction accuracy and monotonicity with respect to human opinion scores. It presented a detailed evaluation of several IQA models, including RIQMC, QMC, RCIQM (RR methods), as well as NIQMC and MDM (NR methods), on the CID2013 and CCID2014 datasets.

As shown in Table 9, the proposed method achieved state-of-the-art performance, with the highest PLCC and SRCC values on both datasets: 0.9581 / 0.9369 on CID2013 and 0.9286 / 0.9178 on CCID2014. These results confirmed the model's superior ability to accurately predict perceptual image contrast quality in the absence of reference images, highlighting its robustness and generalization capability across different dataset conditions.

**Table 9.** Performance comparison of the proposed NR-IQA model and popular competing indices on two benchmark datasets.

| Reference | Method | PLCC (CID2013) | SRCC (CID2013) | PLCC (CCID2014) | SRCC (CCID2014) |
|---|---|---|---|---|---|
| Gu et al. 2016 [14] | RIQMC (RR) | 0.8619 | 0.8010 | 0.8701 | 0.8430 |
| Liu et al. 2014 [44] | QMC (RR) | 0.7710 | 0.7071 | 0.8960 | 0.8722 |
| Liu et al. 2016 [45] | RCIQM (RR) | 0.8866 | 0.8541 | 0.8845 | 0.8565 |
| Gu et al. 2016 [41] | NIQMC (NR) | 0.7225 | 0.6458 | 0.8438 | 0.8113 |
| Nafchi et al. 2016 [15] | MDM (NR) | 0.9279 | 0.8980 | 0.8719 | 0.8368 |
| **The current study** | **Our proposed model (EfficientNet-B0)** | **0.9581** | **0.9369** | **0.9286** | **0.9178** |

Therefore, we saw augmentation not as a new achievement, but as an essential preprocessing step to support model training in data-limited conditions. Although augmentation was not inherently innovative, it has significantly improved performance by reducing overfitting and improving generalization. Indeed, the significant improvement in model accuracy observed in our experiments highlights the critical role of augmentation in stabilizing learning and increasing predictive robustness.

*4.6. Limitation*

While the proposed method demonstrated promising results in contrast-distorted image quality assessment, several limitations remained. First, the evaluation is conducted primarily on benchmark datasets such as CID2013 and CCID2014, which, despite their utility, did not fully capture the complexity and diversity of real-world distortions. Second, although data augmentation was employed to alleviate overfitting, it did not substitute for the availability of large-scale, high-quality annotated datasets. Lastly, cross-dataset generalization and performance in real-world deployment scenarios were not explicitly examined, suggesting the need for further validation in diverse imaging conditions.

## 5. Discussion

The main goal of our work was to develop an effective and efficient framework for no-reference image contrast assessment, and our findings suggested that leveraging pre-trained architectures like EfficientNet-B0, when appropriately customized, offered a promising route. Based on the results presented, our EfficientNet-B0-based model generally outperformed both MobileNetV2 and ResNet18 in Pearson $\rho$ and Spearman $\rho$ and our defined adjusted accuracy. It indicated that this architectural efficiency translated well to the nuances of contrast quality perception.

It was interesting to note the trade-off with training time. MobileNetV2, true to its design for resource-constrained environments, trained the fastest. This highlights an important consideration for practical applications: if sheer speed of deployment or retraining was the absolute priority, MobileNetV2 presented a compelling alternative, even if there was a slight hit in predictive accuracy compared to EfficientNet-B0. ResNet18, while a stalwart architecture, didn't quite match EfficientNet-B0 in accuracy or MobileNetV2 in speed in our specific contrast assessment task.

We used Siamese networks for image contrast assessment in another evaluation, and while Siamese networks are effective for learning pairwise similarity, their performance is limited for fine-grained image contrast assessment. Joint weighting and contrast loss designs often underrepresent subtle changes in luminance and gradient that are critical for perceptual quality. Cross-domain generalization is poor, and performance degrades on datasets with different image statistics. Conventional backbones may lack sufficient background aggregation to capture complex texture-structure relationships, and standard loss functions may not be perceptually aligned.

When we looked at the broader landscape, the aim was to see if such a customized EfficientNet-B0 could indeed hold its own. While the comparisons were against other deep learning models, our initial motivation (and one of our stated contributions) was to offer an improvement over both classical models and more general deep IQA models that did not specifically target contrast. The high correlation scores from EfficientNet-B0 suggested that we were on the right track in that regard, though a more extensive benchmark against a wider array of those classical and general IQA methods would be needed to fully substantiate that claim across diverse datasets. Of course, we should acknowledge some limitations. Our experiments were primarily conducted on the CCID2014 and CID2013 datasets. While it was a good starting point for contrast changes, the real world threw a much wider variety of content and distortion types at us. Testing this framework on a broader range of IQA datasets, especially those with diverse contrast characteristics or other co-occurring distortions, would be a crucial next step. It would also be interesting to experiment with other members of the EfficientNet-B0 family (e.g., B0-B7) to see if a slightly larger model offered a worthwhile accuracy boost without too much of a speed penalty. Further refinement of the hybrid loss function, perhaps by exploring different ways to sample or weight the ranking pairs, could also yield improvements. And finally, a more direct and extensive comparison with leading hand-crafted contrast models would really help solidify the practical advantages of this learned approach.

## 6. Conclusion and feature Work

In this study, we addressed the challenge of no-reference image contrast quality assessment by customizing EfficientNet-B0 with a regression head tailored for perceptual prediction. Experimental results on the CID2013 and CCID2014 datasets demonstrated that the proposed framework achieved strong alignment with subjective human ratings, surpassing conventional deep backbones such as ResNet-18 and MobileNetV2. These findings reinforce the notion that lightweight yet expressive transfer learning architectures can effectively capture perceptual attributes like contrast distortion, which have historically been underrepresented in IQA research.

Despite these promising results, several limitations remain. The reliance on synthetic and benchmark datasets, while useful for controlled evaluation, does not fully represent the diversity of real-world distortions. Moreover, although data augmentation improved generalization, the lack of large-scale contrast-specific annotated datasets still constrains model robustness. Recent systematic reviews highlight the importance of validating IQA models in clinical and cross-domain scenarios, where subtle contrast variations directly influence diagnostic or operational decisions.

Looking forward, several directions emerge for advancing this work. First, broader cross-dataset validation, including authentic and multimodal data (e.g., medical, low-light, and autonomous driving images), is essential to establish real-world generalizability. Second, exploring larger EfficientNet variants (B1–B7) or hybrid ensembles could balance accuracy and efficiency while preserving scalability. Third, refinement of the hybrid loss function through perceptually inspired objectives, such as those grounded in visual attention mechanisms or contrast masking models, could enhance sensitivity to subtle quality degradations.

Finally, integrating this framework into applied settings, such as radiology or streaming media, would not only benchmark practical utility but also accelerate the adoption of NR-IQA methods in resource-constrained environments Overall, this work provides a foundation for future research on contrast-aware IQA, highlighting the potential of lightweight pre-trained networks in bridging the gap between computational efficiency and perceptual fidelity. By expanding datasets, refining architectures, and aligning training objectives with human visual perception, subsequent efforts can further solidify the role of deep learning in automated and scalable image quality assessment.

**Author contributions:**

JHJ conducted the primary experiments, wrote the manuscript, and prepared the figures. BM contributed to writing the manuscript, while OZ reviewed it and provided critical feedback. EA, RA, and HM contributed to manuscript revision and interpretation of results. All authors — JHJ, BM, OZ, EA, RA, and HM — have read and approved the final version of the manuscript.

**Funding:**
This research received no external funding.

**Data availability:**
The experiments reported in this study were conducted using two publicly accessible datasets: the Contrast Changed Image Database 2014 (CCID2014) [14] and the Camera Image Database 2013 (CID2013) [43].

**Declarations**
**Conflict of interest**:
The authors declare no conflicts of interest.